\documentclass[lettersize,journal]{IEEEtran}

\usepackage{amsmath,amsfonts}
\usepackage{algorithmic}
\usepackage{algorithm}
\usepackage{array}
\usepackage[caption=false,font=normalsize,labelfont=sf,textfont=sf]{subfig}
\usepackage{textcomp}
\usepackage{stfloats}
\usepackage{url}
\usepackage{verbatim}
\usepackage{graphicx}
\usepackage{cite}

\usepackage[table]{xcolor} 
\usepackage{multirow}

\begin{document}

\title{DCP-CLIP:A Coarse-to-Fine Framework for Open-Vocabulary Semantic Segmentation with Dual Interaction}

\author{Jing Wang, Huimin Shi, Quan Zhou,~\IEEEmembership{Senior Member,~IEEE}, Qibo Liu, Suofei Zhang, Huimin Lu,~\IEEEmembership{Senior Member,~IEEE}

\thanks{Manuscript received XXXX XX, 2025; revised XXXX XX, 2025; accepted XXXX XX, 2025. This work was jointly supported in part by the National Natural Science Foundation of China under Grants 62476139, the Natural Science Foundation of Jiangsu Province under Grants BK2024023.}

\thanks{\emph{Corresponding author: Quan Zhou}}

\thanks{Jing Wang, Huimin Shi, Quan Zhou, and Qibo Liu are with National Engineering Research Center of Communications and Networking, Nanjing University of Posts \& Telecommunications, Nanjing 210003, P. R. China. Quan Zhou is also with the Institute for Advanced Ocean Research (Nantong), Southeast University, Nantong 226334, P. R. China. (e-mail: quan.zhou@njupt.edu.cn).}
\thanks{Suofei Zhang is with the Department of Internet of Things, Nanjing University
of Posts \& Telecommunications, Nanjing 210003, P. R. China (e-mail: zhangsuofei@
njupt.edu.cn).}
\thanks{Huimin Lu is with the School of Automation, Southeast University, Nanjing
210096, P. R. China, and the Institute for Advanced Ocean Research (Nantong), Southeast University, Nantong 226334, P. R. China. (e-mail: dr.huimin.lu@ieee.org).}
}
\markboth{Journal of \LaTeX\ Class Files,~Vol.~14, No.~8, August~2021}%
{Shell \MakeLowercase{\textit{et al.}}: A Sample Article Using IEEEtran.cls for IEEE Journals}

\maketitle

\begin{abstract}

The recent years have witnessed the remarkable development for open-vocabulary semantic segmentation (OVSS) using visual-language foundation models, yet still suffer from following fundamental challenges:
(1) insufficient cross-modal communications between textual and visual spaces, and (2) significant computational costs from the interactions with massive number of categories. To address these issues, this paper describes a novel coarse-to-fine framework, called DCP-CLIP, for OVSS. Unlike prior efforts that mainly relied on pre-established category content and the inherent spatial-class interaction capability of CLIP, we dynamic constructing category-relevant textual features and explicitly models dual interactions between spatial image features and textual class semantics. Specifically, we first leverage CLIP’s open-vocabulary recognition capability to identify semantic categories relevant to the image context, upon which we dynamically generate corresponding textual features to serve as initial textual guidance. Subsequently, we conduct a coarse segmentation by cross-modally integrating semantic information from textual guidance into the visual representations and achieve refined segmentation by integrating spatially enriched features from the encoder to recover fine-grained details and enhance spatial resolution. In final, we leverage spatial information from the segmentation side to refine category predictions for each mask, facilitating more precise semantic labeling. Experiments on multiple OVSS benchmarks demonstrate that DCP-CLIP outperforms existing methods by delivering both higher accuracy and greater efficiency.
\end{abstract}

\begin{IEEEkeywords}
Open-vocabulary semantic segmentation, Cross-modal alignment, Dynamic category construction.
\end{IEEEkeywords}

\section{Introduction}
\IEEEPARstart{S}{emantic} segmentation is a fundamental task in computer vision that aims to assign a semantic label to each pixel in an image. Traditional semantic segmentation methods \cite{Alpher01,Alpher02,Alpher03,Alpher04} generally assume a fixed set of categories during training and inference, limiting generalization to unseen classes. In contrast, open-vocabulary semantic segmentation \cite{Alpher05,Alpher06,Alpher07} has emerged to address this limitation by enabling models to segment pixels based on arbitrary, text-defined categories.

\begin{figure}[t]
    \centering
    \includegraphics[width=\columnwidth]{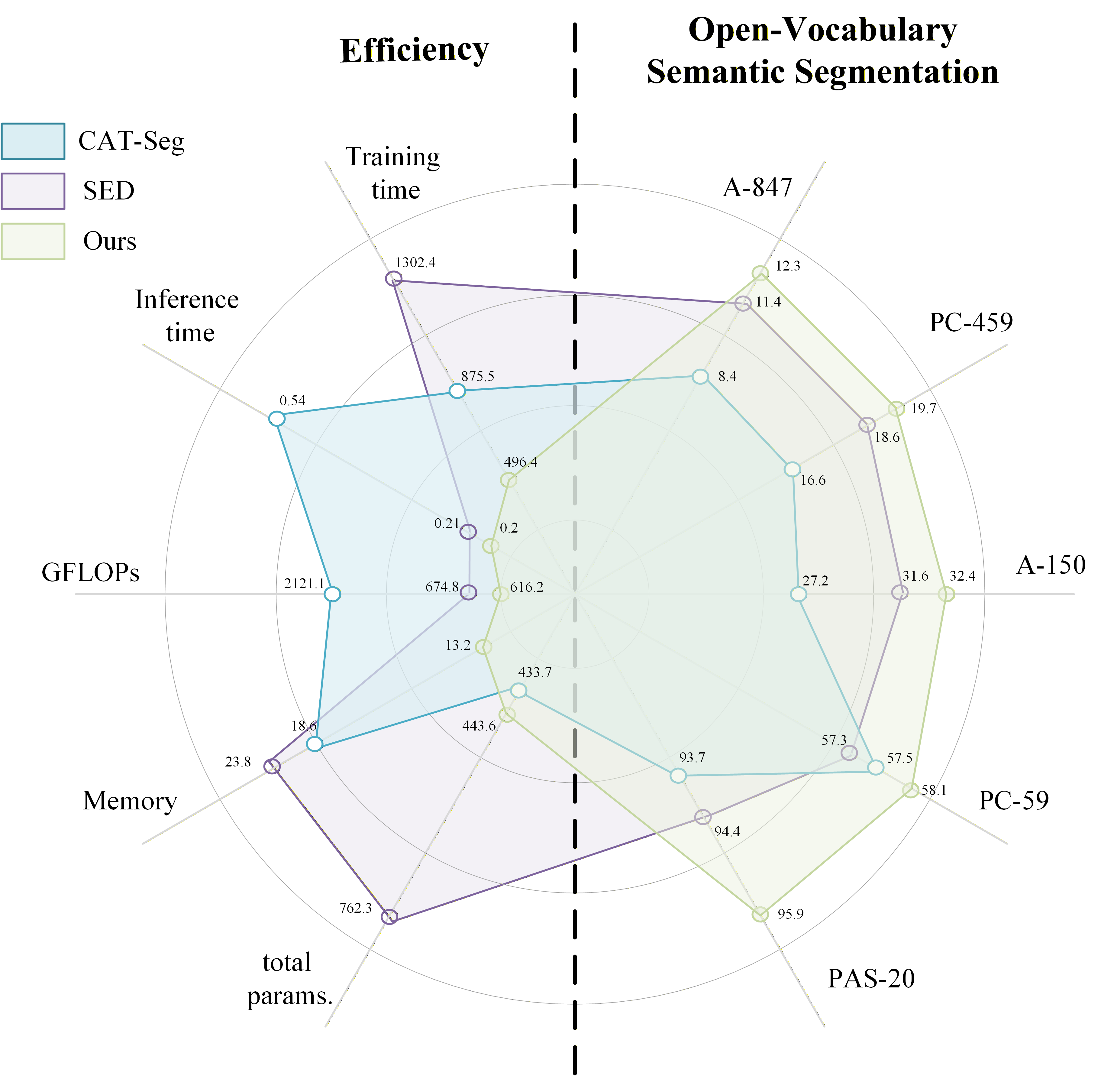}
    \caption{\textbf{Comparison of efficiency and open-vocabulary semantic segmentation performance among CAT-Seg (blue), SED (purple), and our method (green).} The left side shows efficiency metrics, and the right side shows segmentation accuracy. GFLOPs and inference time are measured on a single NVIDIA 3090 GPU. Our method achieves superior segmentation accuracy with balanced efficiency.}
\end{figure}

The development of deep learning (DL) has driven significant efforts to address this more challenging setting. Early works \cite{Alpher05,Alpher06,Alpher07,xu2022groupvit} target the construction of a unified embedding space for effective correspondence between visual features and category semantics. These methods either align visual features with predefined category embeddings\cite{Alpher05,Alpher06}, or directly embed the natural language information as queries to search the desired targets in the pixel features\cite{xu2022groupvit}. However, they struggle to accurately associate visual regions with semantic concepts due to the inherent modality gap between pixel-level visual features and high-level textual category representations.

\begin{figure*}[ht]
    \centering
    \includegraphics[width=0.78\linewidth]{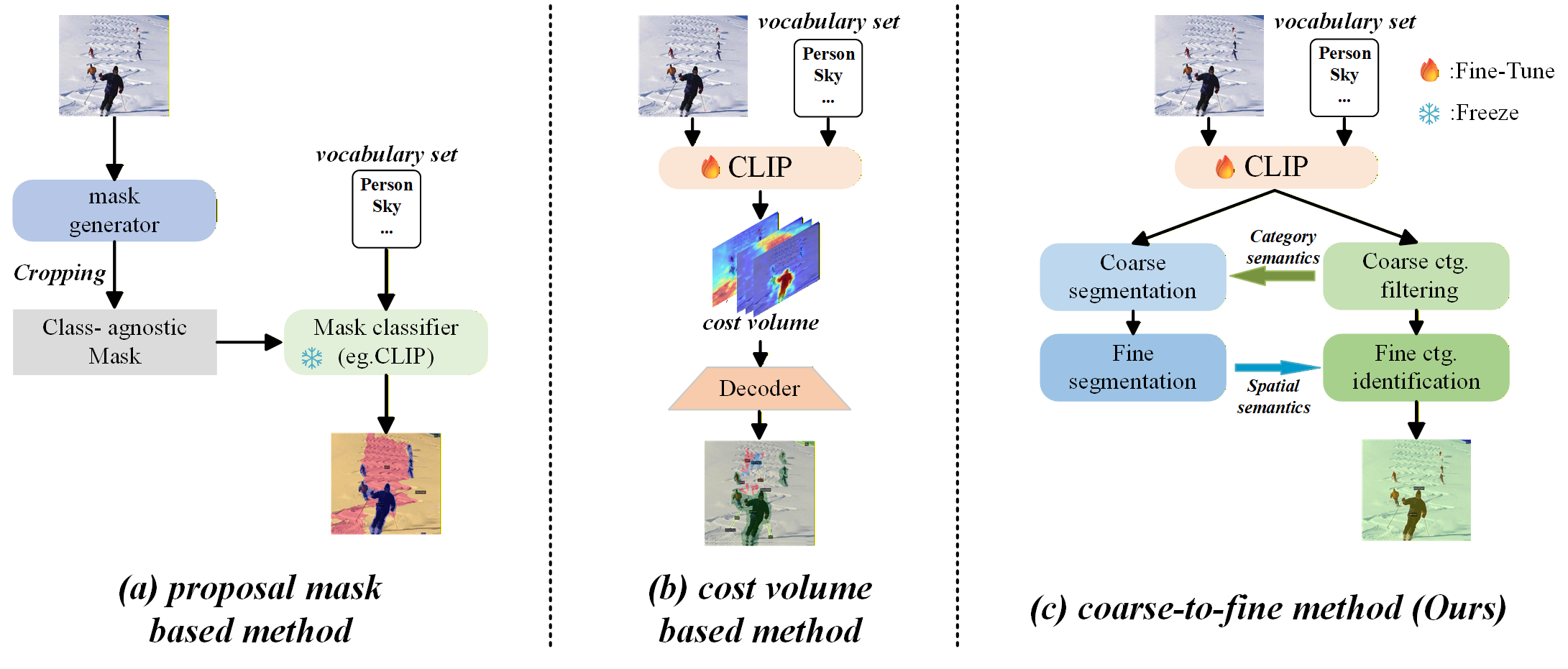}
    \caption{\textbf{Illustration of two representative OVSS methods and our approach.} (a) Proposal mask-based method. (b) Cost volume-based method. (c) Our DCP-CLIP, which adopts a coarse-to-fine framework and introduces a dual interaction mechanism to explicitly enhance semantic alignment between textual and visual spaces.}
    \label{fig1}
\end{figure*}

Recently, vision-language models such as CLIP \cite{Alpher08} and ALIGN \cite{Alpher25} learn aligned image-text feature representations from billion-scale image-text pairs, exhibiting superior open-vocabulary generalization ability in classification tasks. Motivated by this, several works \cite{Alpher09,Alpher13,Alpher12,Alpher15} have explored their application in open-vocabulary semantic segmentation. Early efforts typically adopt a two-stage framework \cite{Alpher10,Alpher11}, proposing proposal mask based method as shown in Fig.~\ref{fig1} (a), where class-agnostic mask regions are first generated and then classified using pre-trained vision-language models. However, training an independent segmentation network on a closed set introduces high training costs and limits scalability to open-vocabulary settings. Furthermore, the absence of contextual information in the classification process leads to limitation in performance.

Unlike the two-stage approach, the single-stage approach exploits the potential of vision-language pre-trained models to enable fine-grained semantic understanding in open-vocabulary semantic segmentation. Pioneering studies, such as SAN \cite{Alpher17} and FC-CLIP \cite{Alpher18} introduce a side adapter network to predict masks using CLIP embeddings and employ an additional decoder in conjunction with the frozen convolutional CLIP to generate class-agnostic masks, respectively. More recently, CAT-SEG \cite{Alpher19} proposes an alternative method as depicted in Fig.~\ref{fig1} (b) by generating cost volumes and refining them through spatial and class aggregation. Building upon this cost volume based method, SED \cite{Alpher20} integrates a hierarchical convolutional encoder to embed spatial information into the cost maps, while adopting skip-layer fusion in the decoder to enhance feature integration and representation. 


However, most of them primarily rely on cross-modal alignment of CLIP model pre-trained on image-text classification, which may be disrupted during fine-tuning for segmentation tasks as discussed in \cite{Alpher18, xu2023maskclip++}. Moreover, previous methods typically perform an exhaustive traversal over all categories in the open vocabulary, which often leads to high computational costs.

To address these issues, we propose DCP-CLIP, a novel coarse-to-fine framework with dual interaction that explicitly enhances communications between textual and visual spaces, as illustrated in Fig.~\ref{fig1} (c). Differing from prior approaches that primarily rely on pre-established category content, we dynamically constructs category-relevant textual embedding based on coarse category filtering. The interaction from the textual space to the visual space is established in the coarse segmentation stage, where we inject high-level semantics from textual embedding into the visual representations through cross-modal integration. To further enhance the segmentation quality, we introduce spatially enriched visual features extracted from the encoder, which are fused into the decoding process to recover fine-grained details and improve spatial resolution. After obtaining the fine-grained segmentation results, we leverage the spatial distribution of the predicted masks to refine the category assignments for each region, thereby completing the reverse interaction from the visual space back to the textual space. Extensive experiments on multiple open-vocabulary semantic segmentation benchmarks demonstrate the effectiveness of our proposed methods in terms of both accuracy and efficiency.


Our main contributions can be summarized as follows:
\begin{itemize}
\item We propose DCP-CLIP, a novel coarse-to-fine framework with dual interaction that explicitly enhances semantic communication between textual and visual spaces and supports more powerful open-vocabulary segmentation capabilities.
\item Instead of exhaustively traversing the all categories, we propose a dynamical category select mechanism to select categories relevant to the image context. It not only significantly improves inference speed, but also enhances training efficiency by reducing memory consumption and training time.
\item We evaluated our method on multiple open-vocabulary semantic segmentation datasets. The extensive experimental results demonstrate that our method outperforms state-of-the-art approaches, offering both rapid inference and superior precision, while requiring only minimal training cost.
\end{itemize}

\section{Related Works}
In this section, we begin by reviewing the foundations of pre-trained vision-language models, which have profoundly shaped recent developments in open-vocabulary semantic segmentation (OVSS). We then discuss recent progress in tuning strategies for vision-language models, highlighting techniques that adapt these powerful models to downstream tasks with minimal supervision. Finally, we provide a comprehensive overview of the evolution of OVSS research, examining both early approaches and recent advancements that leverage large-scale vision-language models.

\subsection{Vision-language models}
vision-language models\cite{Alpher21,Alpher22,Alpher23,Alpher08} have significantly advanced the field of multi-modal learning by establishing a unified semantic space for both visual and textual modalities. By jointly learning visual representations and language semantics, vision-language models effectively bridge the gap between pixels and words, making them powerful foundations for tasks like image captioning, visual question answering, and open-vocabulary recognition. Early research \cite{Alpher21,Alpher22}primarily employed convolutional neural networks (CNNs) for visual feature extraction, coupled with recurrent neural networks (RNNs) or long short-term memory (LSTM) networks for generating text descriptions. Motivated by the remarkable achievements of Transformers in natural language processing, ViLBERT ‌\cite{Alpher23} implement a Bert-based dual-flow architecture that processes image and text information separately and fuses it through an interaction layer, which performs well in visual and linguistic tasks. With the advancement of pre-training techniques, vision-language models pre-trained on large-scale data draw significant attention. For instance, CLIP\cite{Alpher08}, pre-trained on 400 million image-text pairs by contrastive learning approach, has shown remarkable performance in classification tasks. The powerful cross-modal alignment capability of CLIP has advanced the development of various vision downstream tasks, such as zero-shot classification\cite{sammani2024clip_zero_shot}, object detection\cite{zhong2022regionclip}, and image segmentation\cite{Alpher12, Alpher13}. RegionCLIP\cite{zhong2022regionclip} extends CLIP to object detection by aligning region-level visual features with textual category embeddings. MaskCLIP\cite{Alpher13} adapts the CLIP model for segmentation by removing global pooling and enabling dense prediction tasks through contrastive learning with masked region features. The emergence of pre-trained vision-language models enables unified processing of visual and linguistic information, unlocking broad applicability across diverse multi-modal tasks. In this paper, we also build upon CLIP’s strong multi-modal foundation, and further incorporate multi-scale visual information to obtain semantically rich fused features.

\subsection{Tuning of vision-language models }
In order to fully unleash the powerful capabilities of large-scale vision-language models across various downstream tasks, a variety of adaptation strategies\cite{Alpher27,Alpher30,Alpher29,zhang2020sidetuning} have been explored to better align vision-language models with task-specific objectives. early research like CoOp\cite{Alpher27} and CoCoOp\cite{Alpher26} propose to fine-tune CLIP's text embedding by learning text prompts, allowing the model to learn more context-specific textual representations. Instead, MenglinJia et al. \cite{Alpher28} choose to an alternative approach by adding prompts to the visual encoder for fine-tuning, which even outperforms full fine-tuning in many cases. Motivated by it, CLIP Surgery \cite{,Alpher29} explores CLIP internal architecture and modify the way it computes attention. Moreover, Tip-Adapter\cite{Alpher31} and VL-Adapter\cite{Alpher30} are innovative designs that incorporate trainable adapters to effectively aggregate the features extracted from frozen CLIP. Side-Tuning \cite{zhang2020sidetuning} proposes a simple and efficient network adaptation strategy by introducing a lightweight side network that is additively combined with a pre-trained frozen backbone. While these works have indeed improved the performance of vision-language models on image-level tasks, extending them to dense prediction tasks—such as semantic segmentation—remains a significant challenge. This is primarily due to the lack of fine-grained alignment at the pixel or region level and the difficulty in transferring global semantic representations to spatially structured outputs. Our work explores ways to extend the fine-tuning CLIP beyond its original image and text embedment alignment into pixel-level predictions by introducing dual interactions between spatial and textual spaces.

\subsection{Open-vocabulary semantic segmentation}
Open-vocabulary semantic segmentation (OVSS) aims to segment images into semantic regions from an unlimited categories set. Initially, research \cite{Alpher05,Alpher06,Alpher07} focus on aligning visual features with word embeddings by developing methods for learning feature mapping. With the advent and success of large-scale vision-language model like CLIP\cite{Alpher08} and ALIGN \cite{Alpher25}, OVSS methods \cite{Alpher09,Alpher13,Alpher12,Alpher15} have begun leveraging CLIP\cite{Alpher08} for open-vocabulary semantic segmentation. SimSeg\cite{Alpher39} propose a two-stage framework consisting of a class-agnostic mask generator and a frozen CLIP encoder for class recognition. This proposal mask based method has since become the foundation for many follow-up works. In order to improve the mask classification performance of CLIP, OVSeg\cite{Alpher12} adopt a mask prompt tuning scheme to realize mask adaptive of CLIP. Meanwhile, FreeSeg\cite{Alpher15} abstract multi-granularity concepts into a text representation,realizing generalization to arbitrary text descriptions. On top of leveraging CLIP, several recent works \cite{Alpher14,wang2023diffusion_model_segmenter} have also explored the use of diffusion models ODISE\cite{Alpher14} harnesses the power of Text-to-Image diffusion models\cite{rombach2022high} to effectively generate high-quality mask proposals, which are subsequently refined and classified through discriminative model. DiffSegmenter \cite{wang2023diffusion_model_segmenter} also repurposes text-to-image diffusion models for open-vocabulary semantic segmentation by extracting cross-attention maps as class-aware localization cues.

As fine-tuning strategies for vision-language models continue to evolve, several studies\cite{Alpher20,Alpher17,Alpher16} exploit one-stage methods that directly adapt CLIP for fine-grained semantic understanding in open-vocabulary semantic segmentation. LSeg \cite{Alpher16} align pixel-level features with pre-trained textual embeddings to achieve flexible segmentation across diverse categories. SAN \cite{Alpher17} and ZegCLIP \cite{zhou2023zegclip} improve the segmentation performance of CLIP by introducing a side adapter network and adopting deep prompt fine-tuning, respectively. Recently, CAT-Seg \cite{Alpher19} and SED \cite{Alpher20} attempt to refine the pixel-level cost volumes generated by the fine-tuned CLIP by leveraging spatial-class aggregation and hierarchical feature fusion. Meanwhile, In addition, EBSeg \cite{Alpher32} explores integrating features from SAM \cite{Alpher33} to enrich the spatial information and balance image embeddings between training classes and novel classes. At the architectural level, BBN \cite{pan2025purifythen} proposes a bi-directional bridge network that first purifies noisy textual embeddings and then guides cross-modal alignment.
 
In contrast to prior methods, our approach primarily focuses on addressing two key challenges: (1) reducing the computational cost in the open-vocabulary setting through a pre-filtering method that lowers the complexity over categories; and (2) enhancing the cross-modal alignment by introducing additional vision-language interactions.

\begin{figure*}[ht]
    \centering
    \includegraphics[width=0.95\textwidth]{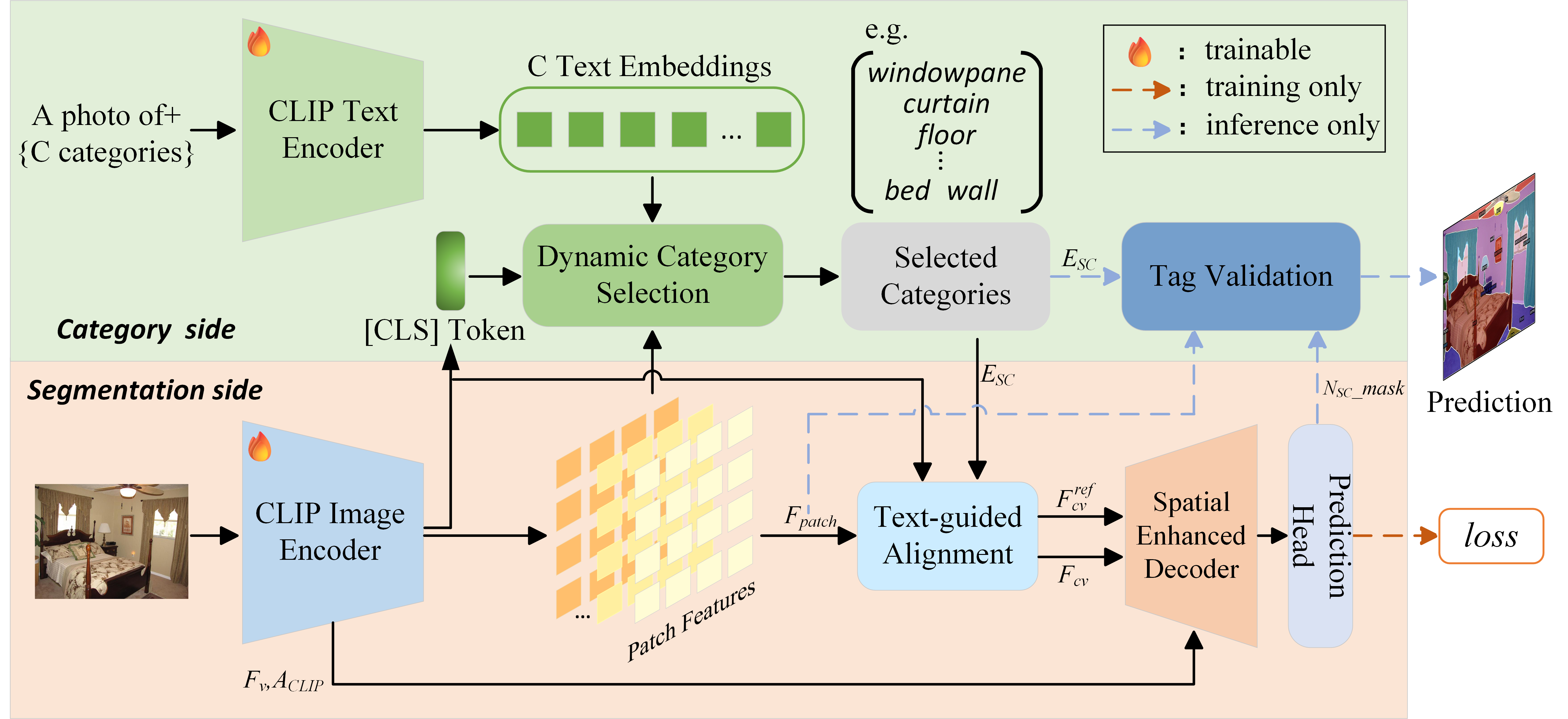}
    \caption{\textbf{Overview of our proposed DCP-CLIP.} The Dynamic Category Selection module adaptively selects relevant categories based on the image content. Next, we perform cross-modal semantic learning under the guidance of the selected categories, followed by fine-segmentation that restores spatial details and resolution through a Spatial Enhanced Decoder. During training, the fine segmentation output is used for supervision. During inference, Tag Validation module leverages spatial context to refine open-vocabulary predictions. }
    \label{fig2}
\end{figure*}

\section{Method}
In Fig.~\ref{fig2}, We present an overview of our proposed open-vocabulary semantic segmentation model, DCP-CLIP, which consists of four main components: (1) Dynamic Category Selection that adaptively selects semantically relevant categories based on global image content. (2) Text-guided alignment that derives dynamic category guidance from selected categories and refines visual features through cross-attention-based cross-modal interaction. (3) Spatial Enhanced Decoder that performs intra-modal alignment to restore fine-grained details and enhance spatial resolution. (4) Tag Validation that leverages spatial context to adjust category predictions. Based on the above overview, we proceed to introduce each component in detail.

\subsection{Dynamic Category Selection}
Segmentation methods based on a predefined set of textual semantic categories have been widely adopted in open-vocabulary semantic segmentation. However, as the number of predefined categories increases, computational efficiency degrades significantly. Not all categories appear in every image; in fact, most images contain only a few semantic categories. Therefore, exhaustively traversing the entire category set not only leads to unnecessary computation but may also introduce semantic noise interference. To improve efficiency and accuracy, we propose a dynamic category selection mechanism that leverages image-level semantic features to select potentially relevant categories.

Given an image $\mathit{I} \in  \mathbb{R^{\mathrm{\mathit{H}\times \mathit{W} \times \mathit{3}} } }$, where $\mathit{H}$ and $\mathit{W}$ respectively represent the length and width of the image. We first use VIT-based CLIP image encoder to extract visual features which comprises [CLS] token ${F_{cls}}\in  \mathbb{R^{\mathrm{ \mathit{1}\times \mathit{D} } } } $ and patch features $\boldsymbol{F_{patch}}\in \mathbb{R^{\mathrm{\mathit{N}\times \mathit{D}} } } $, with $\mathit{D}$ is the feature dimension of CLIP and $\mathit{N}$ is the number of patch tokens. Given an arbitrary set of category names $\left \{ T_{1}, T_{2}...T_{\mathit{C}} \right \} $, we fill it into text template as text input of CLIP text encoder to generate $\mathit{C}$ text embeddings $\boldsymbol{E_{t}}\in  \mathbb{R^{\mathrm{\mathit{C}\times \mathit{D} } }} $. $\mathit{C}$ represents the total number of categories. Then, we fed visual features and text embeddings to Dynamic Category Selection module to select categories that may appear in the image. By computing the cosine similarity \cite{rocco2017convolutional} between patch features and text embeddings, as well as between the [CLS] token and text embeddings in the feature space, we obtain the patch-level classification score $\boldsymbol{S _{patch}}\in \mathbb{R^{\mathrm{\mathit{C}\times \mathit{N}} } }$ and image-level classification score $\boldsymbol{S _{img}}\in \mathbb{R^{\mathrm{\mathit{1}\times \mathit{C} } }}$:
\begin{equation}
\begin{split}
&\boldsymbol{S_{patch}}=\frac{\boldsymbol{F_{patch}}\cdot \boldsymbol{E^{T}_{t}}}{\parallel \boldsymbol{F_{patch}} \parallel\parallel \boldsymbol{E_{t}} \parallel}
\\
\boldsymbol{S _{img}} &= softmax(\tau \cdot\frac{\boldsymbol{F_{cls}}\cdot \boldsymbol{E^{T} _{t}}}{\parallel {\boldsymbol{F_{cls}}} \parallel\parallel \boldsymbol{E_{t}} \parallel} ) )
\end{split}
\end{equation}
where $\tau$ is the logit scale for softmax, $T$ denotes the transpose of a matrix, $softmax(\cdot)$ represents softmax function.

Under the open-set semantic setting, constructing the category set requires balancing the accurate capture of seen categories with the exploration and discovery of unseen ones. Therefore, we adopt a joint construction strategy at both the image and patch levels, Specifically, the image level focuses on capturing dominant categories, whereas the patch level suppresses their influence to explore the potential presence of secondary categories. The detailed description of the selection strategy is presented below.

At the patch level, we employ an image-level-guided method to remove redundant information in patch-level that originates from dominant categories, allowing each patch to focus on identifying the potential presence of other categories. Specifically, the image-level classification scores are broadcast to \( \mathbb{R}^{C \times N} \) and leveraged along the class dimension to guide the suppression of redundant semantic information associated with dominant categories to generate the refined patch-level classification scores \( \boldsymbol{S}^{\mathrm{ref}}_{\mathrm{patch}} \in \mathbb{R}^{C \times N} \):
\begin{equation}
\boldsymbol{S}^{\mathrm{ref}}_{\mathrm{patch}} = \boldsymbol{S}_{\mathrm{patch}} -  \left( \boldsymbol{S}^{\top}_{\mathrm{img}} \cdot \boldsymbol{1}^{\top} \right) \odot \boldsymbol{S}_{\mathrm{patch}} 
\end{equation}

where$\odot$ represents Hadamard product, $T$ denotes the transpose of a matrix.

After eliminating information from the dominant categories, the category set is constructed at the patch level. Specifically, we identify the most relevant category for each token based on the refined classification scores and take the union of these per‐token predictions to form a candidate set $\mathcal{C}_{\text{patch}}$:

\begin{equation}
\mathcal{C}_{\text{patch}} = \bigcup_{i=1}^{N} \{ \hat{c}_i \}, \quad \text{where } \hat{c}_i = \arg\max_{c} \, \boldsymbol{S}^{\mathrm{ref}}_{\mathrm{patch}}(c, i)
\end{equation}

Here, \(\hat{c}_i\) denotes the most relevant category for the \(i\)-th token, obtained by selecting the category \(c\) that maximizes the patch-level category score \(\boldsymbol{S}^{\mathrm{ref}}_{\mathrm{patch}}(c, i)\).

At the image level, we adopt a straightforward approach by applying a confidence threshold $\theta$ and selecting categories whose classification scores exceed this threshold to form a candidate set $\mathcal{C}_{\text{image}}$. The final selected categories set is then obtained by taking the union of the two candidate sets:

\begin{equation}
\mathcal{C}_{\text{image}} = \{ c \mid s^{\text{img}}_c > \theta \}
\end{equation}
\begin{equation}
\mathcal{C}_{\text{final}} = \mathcal{C}_{\text{patch}} \cup \mathcal{C}_{\text{image}}
\end{equation}

where \( s^{\text{img}}_c \) denotes the image-level classification score for class \( c \) , and \( \mathcal{C}_{\text{final}} \) is the intersection of patch-level and image-level candidate sets.

\subsection{Text-guided alignment}
\begin{figure}[t]
    \centering
    \ \includegraphics[width=\columnwidth]{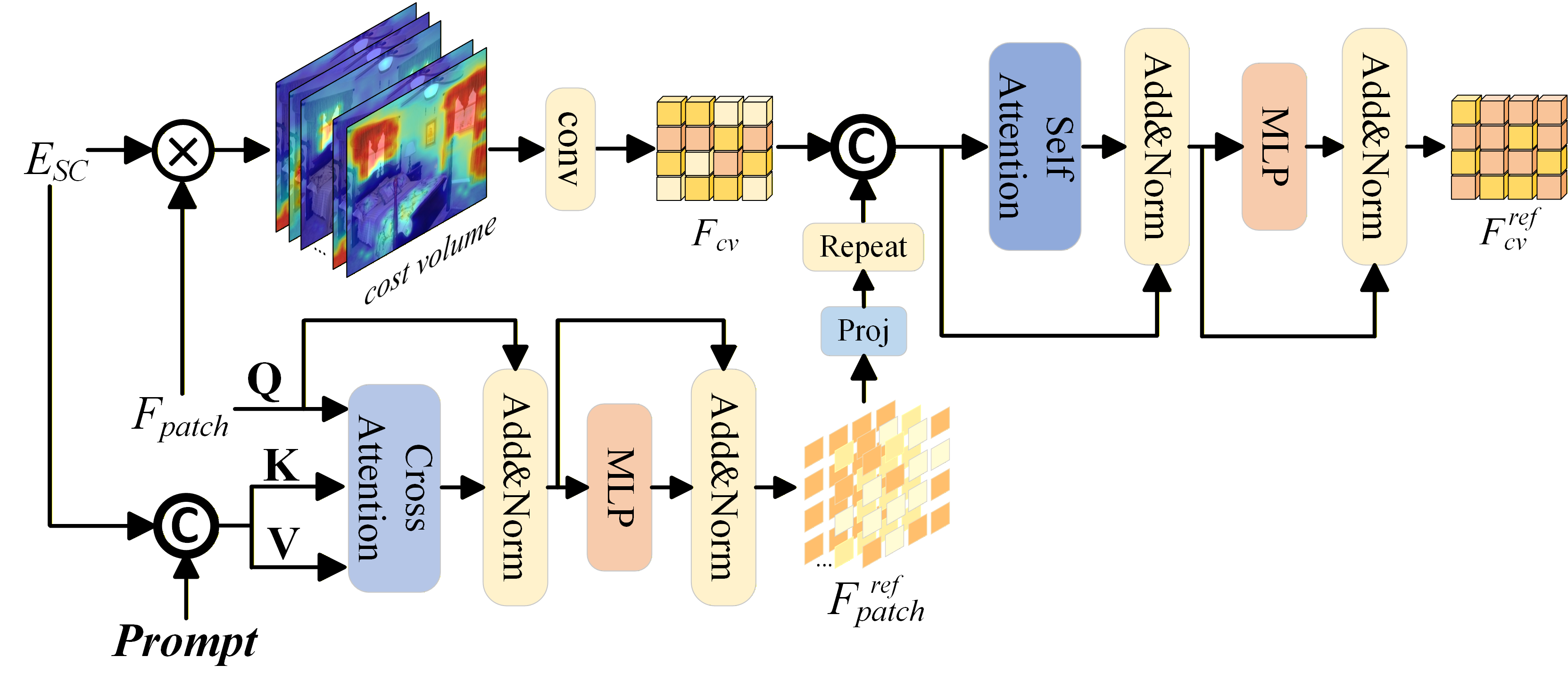}
    \caption{\textbf{The detailed architecture of text-guided alignment.} High-level semantics from text are injected via image features as a bridge, and then fused with initial cost volume features through self-attention to produce more expressive representations.}
    \label{fig3}
\end{figure}
Due to the image-level supervision paradigm adopted during CLIP’s pre-training, directly applying the embeddings of the selected categories to fine-grained tasks like segmentation leads to suboptimal results. To facilitate the fusion of visual and linguistic semantics, we build upon the construction of the cost volume and employ image features as a bridge to incorporate high-level textual information through cross-attention, inspired by \cite{zhang2022segvit,zhou2023zegclip}. The overall structure of our proposed text-guided alignment module is illustrated in Fig.~\ref{fig3}.

Firstly, we perform the cosine similarity computation\cite{rocco2017convolutional} between patch features and selected category embeddings to obtain the initial cost volume. The initial cost volume goes through a convolutional layer to generate the cost volume feature $\boldsymbol{F_{cv}}\in  \mathbb{R^{\mathrm{C_{s}\times N\times D_{c}} }}$, where \( C_s \) denotes the number of selected categories and \( D_c \) denotes the dimension of cost volume feature:
\begin{equation}
\boldsymbol{F_{cv}}=conv(\frac{\boldsymbol{F_{patch}}\cdot \boldsymbol{E _{SC}}}{\parallel \boldsymbol{F_{patch}} \parallel\parallel \boldsymbol{E_{SC}} \parallel})
\end{equation}

To incorporate high-level semantic features from the textual modality, we introduce a parallel branch where image features serve as a bridge to inject enhanced textual semantics into the visual representation. In order to further enhance the high-level semantic properties of the embeddings from textual spaces and better align them with specific image content, We adjust the selected category embeddings \( \boldsymbol{E}_{\text{SC}} \in \mathbb{R}^{C_s \times D} \) in the textual space by concatenating a learnable prompt \( \boldsymbol{P} \in \mathbb{R}^{C_s \times D} \) and the classification logits derived from the [CLS] token. A learnable linear projection \( \phi \) is then applied to map the concatenated representation into the dimensionally aligned textual guidance \( \boldsymbol{T}_{g} \in  \mathbb{R^{\mathrm{C_{s}\times D} } }\), as formulated below:

\begin{equation}
\boldsymbol{T _{g}}  =  \phi(concat[\boldsymbol{F_{cls}}\cdot \boldsymbol{E _{SC}},\boldsymbol{E_{SC}},\boldsymbol{P}])
\end{equation}

Regarding the visual space, the patch features are transformed into queries (\( \boldsymbol{Q} \)) via a linear transformations \( \boldsymbol{W}^{q} \in \mathbb{R}^{D \times D} \). Meanwhile, the textual guidance is independently projected into keys (\( \boldsymbol{K} \)) and values (\( \boldsymbol{V} \)) using separate linear transformations \( \boldsymbol{W}^{k} \in \mathbb{R}^{D \times D} \) and \( \boldsymbol{W}^{v} \in \mathbb{R}^{D \times D} \), respectively.
\begin{equation}
\begin{split}
\boldsymbol{Q} &= \boldsymbol{F_{patch}W^{q}}\in  \mathbb{R^{\mathrm{N\times D} } },
\\
\boldsymbol{K} = \boldsymbol{T_{g}W^{k}}&\in  \mathbb{R^{\mathrm{C_{s}\times D} } },
\boldsymbol{V} = \boldsymbol{T_{g}W^{v}}\in  \mathbb{R^{\mathrm{C_{s}\times D} } }
\end{split}
\end{equation}

After that, the cross-modal learning is implemented via cross-attention. The attention weights are computed by the scaled dot-product similarity between \( \boldsymbol{Q} \) and \( \boldsymbol{K} \), followed by a softmax operation, which is then used to weight the values \( \boldsymbol{V} \). The resulting representations are further processed through residual connections, layer normalization, and a feed-forward network (MLP). Through \( N_{c} \) effective cross-modal interactions between the text and image domains, a refined and semantically enriched patch feature \( \boldsymbol{F}^{\text{ref}}_{\text{patch}} \in \mathbb{R}^{N \times D}\) is obtained:
\begin{equation}
\boldsymbol{Z} = \text{softmax} \left( \frac{\boldsymbol{Q} \boldsymbol{K}^\top}{\sqrt{D}} \right) \boldsymbol{V}, \quad
\boldsymbol{X} = \text{LN} \left( \boldsymbol{Z} + \boldsymbol{F}_{\text{patch}} \right)
\end{equation}
\begin{equation}
\boldsymbol{F}^{\text{ref}}_{\text{patch}} = \text{LN} \left( \text{MLP} ( \boldsymbol{X} ) + \boldsymbol{X} \right)
\end{equation}

To enable deep interaction and context-aware fusion between the cost volume feature and semantically enriched patch feature, We first project and repeat refined patch features $\boldsymbol{F^{ref}_{patch}}$ to match the dimensions and channel of $F_{\mathrm{cv}}$, and then concatenate them together to generate the concatenated features $F_{\mathrm{concat}}\in \mathbb{R}^{C_s \times N\times 2D_c}$. Then, the concatenated features are projected into query, key, and value representations through three linear transformations:
\begin{equation}
\boldsymbol{F_{\mathrm{concat}}}=concat(\boldsymbol{F_{cv}},\phi(repeat(\boldsymbol{F^{ref}_{patch}})))
\end{equation}
\begin{equation}
\boldsymbol{Q} =  \boldsymbol{F_{\mathrm{concat}}}W^q,
\boldsymbol{K} =  \boldsymbol{F_{\mathrm{concat}}}W^k,
\boldsymbol{V} =  \boldsymbol{F_{\mathrm{concat}}}W^v
\end{equation}

where $\{\boldsymbol{W^{q}},\boldsymbol{W^{k}},\boldsymbol{W^{v}}\}\in \mathbb{R}^{2D_c \times D_c}$are learnable weight matrices and \( \phi (\cdot)\) denotes the linear projection.

After that, the attention weights are computed using the scaled dot-product similarity between the queries and keys, followed by a softmax operation. These weights are then used to aggregate the value representations. The resulting features are further refined through residual connections, layer normalization, and a feed-forward network (MLP). Since the detailed computation is similar to Equations~(8) and (9), we omit it here for brevity. By modeling contextual dependencies among image regions through self-attention, the network captures long-range relationships and enhances semantic consistency within the cost volume features, leading to more expressive representations $\boldsymbol{F^{ref}_{cv}}\in \mathbb{R}^{C_s \times N\times D_c}$.
\subsection{Spatial Enhancement Decoder}
Semantic segmentation tasks require high-resolution feature to accurately capture fine-grained spatial details. However, the Vision Transformer architecture in CLIP is primarily optimized for low-resolution image-level classification, making it inherently deficient in preserving the high-resolution and localized features crucial for dense prediction tasks such as semantic segmentation. To address this limitation, We adopt swin transformer to compensate for the spatial information deficiency, and further incorporate a simple U-Net-like upsampling decoder to recover the resolution.

\begin{figure}[t]
    \centering
    \ \includegraphics[width=\columnwidth]{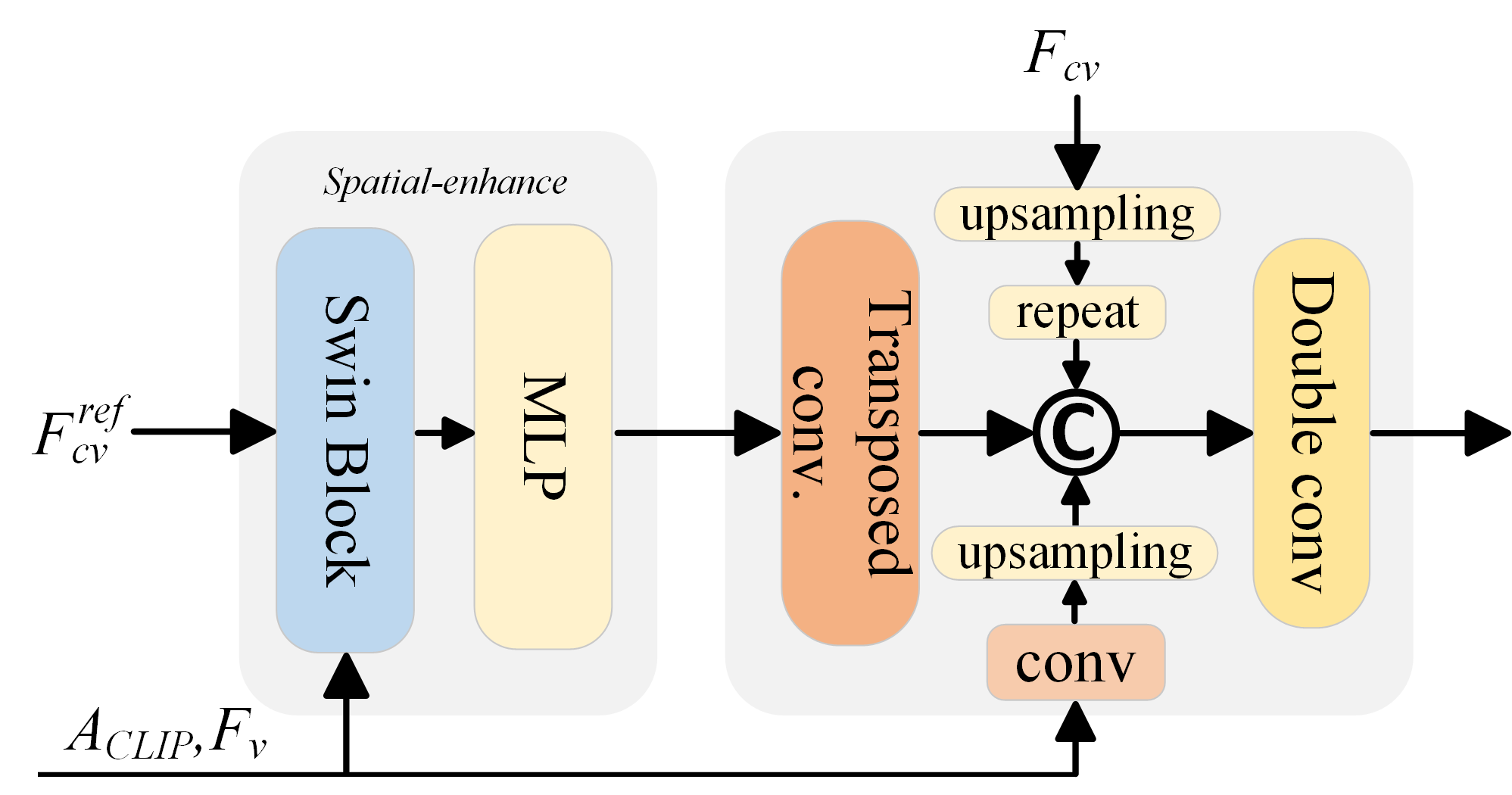}
    \caption{\textbf{The detailed architecture of Spatial Enhanced Decoder.} Spatial enhancement uses a Swin Transformer to model local context with shallow features and attention maps. Deconvolution and multi-source fusion produce refined semantic maps.}
    \label{fig4}
\end{figure}

Fig. ~\ref{fig4} presents the detailed structure of layer of Spatial Enhanced Decoder that comprise with spatial-enhance layers and upsampling decoder layers. To incorporate the multi-scale features from the encoder that encode rich local information, we first apply spatial enhance to capture contextual relationships among neighboring regions. As discussed in \cite{xu2022multi,gao2021tscam}, the inherent self-attention mechanism of Vision Transformers (ViTs) captures pairwise affinities between patches. Therefore, we further integrate attention weights on the basis of shallow feature fusion to enhance the representational capacity of the fused features. Specifically, we perform spatial enhancement using the Swin Transformer~\cite{liu2021swin} to obtain low resolution feature $\boldsymbol{F_{l}}\in \mathbb{R}^{C_s \times N\times D_c}$, with a modified attention mechanism. We concatenate the refined cost volume feature map $\boldsymbol{F^{\mathrm{ref}}_{\mathrm{cv}}}$ with shallow encoder features and attention weights, and apply two linear transformations to generate the query and key representations. Meanwhile, the value representation is obtained by applying a separate linear transformation to $\boldsymbol{F^{\mathrm{ref}}_{\mathrm{cv}}}$. The rest of the Swin Transformer’s architecture and operations remain unchanged. The computation process is as follows:
\begin{equation}
\boldsymbol{F_{\mathrm{m}}}=concat(\boldsymbol{F^{ref}_{cv}},conv(\boldsymbol{F_{v}}),conv\boldsymbol{(A_{CLIP}}))
\end{equation}
\begin{equation}
\boldsymbol{Q} =  \boldsymbol{F_{\mathrm{m}}}W^q,
\boldsymbol{K} =  \boldsymbol{F_{\mathrm{m}}}W^k,
\boldsymbol{V} =  \boldsymbol{F^{ref}_{cv}}W^v
\end{equation}

where $\{\boldsymbol{W^{q}},\boldsymbol{W^{k}}\}\in \mathbb{R}^{3D_c \times D_c}$ and $\{\boldsymbol{W^{v}}\}\in \mathbb{R}^{D_c \times D_c}$are learnable weight matrices. It is worth noting that, as the resolution increases in the later layers of the decoder, we additionally apply an upsampling operation to $\boldsymbol{F_{v}}$ and $\boldsymbol{A_{CLIP}}$ before further Concatenation.

Subsequently, we restore the spatial resolution of the image and integrate spatial information through the decoder of a U-Net-like architecture, which employs transposed convolution layers to upsample feature maps, followed by $3 \times 3$ convolutional layers for progressive spatial resolution recovery. During this process, we concatenate the initial cost map and shallow encoder features with the upsampled feature maps to incorporate additional spatial and semantic information, which can be formulated as:
\begin{equation}
\boldsymbol{F_{h}}= TransConv(\boldsymbol{F_{l}})
\end{equation}
\begin{equation}
\boldsymbol{F_{out}}=concat(\boldsymbol{F_{h}},up(\boldsymbol{F_{cv}}\otimes \mathbf{1}_c),up(\boldsymbol{conv(F_{v}})))
\end{equation}

where $\text{TransConv}(\cdot)$ denotes a transposed convolution operation, and $\text{up}(\cdot)$ denotes an upsampling operation. The symbol $\otimes$ represents a repetition operation, and $\mathbf{1}_c \in \mathbb{R}^{C_s}$ denotes an all-one vector used to repeat the feature along a specific dimension.

We pass the features through a total of $N_s$ layers of the Spatial Enhanced Decoder, ultimately producing a high-resolution and spatially detailed representation $\boldsymbol{F}_{\mathrm{last}}\in \mathbb{R}^{C_s \times (4^{N_s}*N)\times (D_c/2^{N_s})}$. This process effectively bridges the gap between low-resolution transformer features and the high-resolution requirements of semantic segmentation by integrating hierarchical spatial enhancement and progressive upsampling, resulting in more precise and semantically rich dense predictions.

\subsection{Tag Validation}
In this section, we will provide a detailed description of how to leverage existing spatial information to refine the preliminary classification results, thereby achieving more accurate category recognition. Our re-validation module, inspired by the approach like SCAN\cite{liu2024open} in two-stage methods, performing a secondary classification for regions that were initially assigned to unseen categories and maintain the prediction of seen categories.

During inference, we adopt different strategies for seen and unseen categories. For seen categories, we rely on the model’s initial prediction, assuming that it is sufficiently reliable. In contrast, for unseen categories, we apply a tag validation mechanism: we firstly perform a pooling operation over the corresponding segmented region and recompute the similarity with textual embeddings to obtain an updated category prediction. Then, the updated prediction is compared with the original label, and if the original label is not ranked among the top-k most similar categories, it is replaced with the newly inferred category. This re-validation strategy enables the model to correct early-stage misclassifications by exploiting spatially aligned visual information. The specific formulation is as follows:

\begin{equation}
\hat{c}_{mask} = 
\begin{cases}
c_{mask}, & \text{if } c_{mask} \in C_{seen} \\[6pt]
\hat{c}_{mask}^{\text{updated}}, & \text{if } c_{mask} \notin C_{seen} \text{ and rank} > k \\[6pt]
c_{mask}, & \text{otherwise}
\end{cases}
\end{equation}

where $c_{mask}$ denotes the initial category prediction for the masked region, $\hat{c}_{mask}^{\mathrm{updated}}$ is the updated category prediction after re-validation, $C_{seen}$ represents the set of seen categories.

As a result, it effectively enhances the reliability of region-level predictions, particularly in ambiguous or cluttered scenes, and contributes to improved overall segmentation accuracy in open-vocabulary settings.

\section{Experiments}

\begin{table*}[t]
\centering
\caption{\textbf{Comparison with State-of-the-Art Methods.} We use mIoU as the evaluation metric. The best-performing results are highlighted in bold, while the second-best are underlined.}
\resizebox{\textwidth}{!}{
    \begin{tabular}{c|c|cc|c|ccccc}
        \hline
        Method  &  Pub.&   {VLM}&     {Additional Backbone}&{Training dataset} &  {A-847} &     {PC-459} &     {A-150} &     {PC-59} &     {PAS-20}\\ 
        \hline
        SPNet\cite{XianCVPR2019b} &CVPR19&  -& ResNet-101 &  PASCAL VOC  & -& -& -& 24.3 &  18.3 \\ 
        ZS3Net\cite{Alpher05}& NeurIPS19&  -& ResNet-101 &  PASCAL VOC  & -& -& -& 19.4 &  38.3 \\ 
        LSeg\cite{Alpher16} & ICLR22&  CLIP ViT-B/32&ResNet-101&  PASCAL VOC-15  & -& -& -& - &  47.4 \\ 
        GroupViT\cite{xu2022groupvit}&CVPR22 & ViT-S/16&-&  GCC\cite{sharma-etal-2018-conceptual} +YFCC\cite{thomee2016yfcc100m} & 4.3& 4.9& 10.6&  25.9 &  50.7\\ 
         LSeg+\cite{ghiasi2022scaling}&ECCV22 &  ALIGN &ResNet-101 &  COCO-Stuff  & 2.5& 5.2& 13.0& 36.0 &  - \\ 
        ZegFormer\cite{Ding_2022_CVPR}&CVPR22 & CLIP ViT-B/16&ResNet-101&  COCO-Stuff-156 & 4.9& 9.1& 16.9&  42.8 &  86.2\\ 
        OpenSeg\cite{ghiasi2022scaling} &ECCV22&  ALIGN&ResNet-101&  COCO Panoptic\cite{kirillov2019panoptic}+LOc. Narr.\cite{PontTuset_eccv2020}  & 4.4& 7.9& 17.5& 40.1 &  - \\
        ZSseg\cite{xu2022simple}&ECCV22 & CLIP ViT-B/16&ResNet-101&  COCO-Stuff & 7.0& -& 20.5&  47.7 &  88.4\\ 
        DeOP\cite{Han_2023_ICCV} & ICCV23 &  CLIP ViT-B/16&ResNet-101c&  COCO-Stuff-156  & 7.1& 9.4& 22.9& 48.8 &  91.7 \\
        OVSeg\cite{Alpher12} &CVPR23 &  CLIP ViT-B/16&ResNet-101c&  COCO-Stuff+COCO Caption  & 7.1& 11.0& 24.8& 53.3 &  92.6 \\
        SAN\cite{Alpher17} &CVPR23 &  CLIP ViT-B/16&-&  COCO-Stuff  &10.1& 12.6& 27.5& 53.8 &  94.0 \\ 
        PACL\cite{mukhoti2023open} &CVPR23&  CLIP ViT-B/16&-&  GCC\cite{sharma-etal-2018-conceptual}+YFCC\cite{thomee2016yfcc100m}  &-& -& 31.4& 50.1 &  72.3 \\ 
        CAT-Seg\cite{Alpher19}&CVPR24 &  CLIP ViT-B/16&ResNet-101&  COCO-Stuff  & 8.4& 16.6& 27.2& 57.5 &  93.7 \\
        EBSeg\cite{Alpher32}&CVPR24 &  CLIP ViT-B/16&SAM\cite{Alpher33}&  COCO-Stuff  & 11.1& 17.3& 30.0& 56.7 &  94.6 \\
        SCAN\cite{liu2024open}&CVPR24 &  CLIP ViT-B/16&Swin-B&  COCO-Stuff  & 10.8& 13.2& 30.8& \textbf{58.4} &  \textbf{97.0}\\
        SED\cite{Alpher20}&CVPR24 & ConvNeXt-B&-&  COCO-Stuff  & \underline{11.4}& \underline{18.6}& \underline{31.6}& 57.3 &  94.4 \\
        CEL\cite{dao2024class_enhancement_losses}&TMM25 & CLIP R50 &-&  COCO-Stuff  & 9.7& 12.6& 29.9& 55.6 &  91.8 \\
        \rowcolor{gray!20} 
        \textbf{DCP-CLIP(Ours)}&- & CLIP ViT-B/16&-&  COCO-Stuff   & \textbf{12.3}& \textbf{19.7}& \textbf{32.4}&   \underline{58.1} &  \underline{95.9}\\ 
        \hline
        LSeg\cite{Alpher16}&ICLR22 &  CLIP ViT-B/32&ViT-L/16&  PASCAL VOC-15  & -& -& -& - &  52.3 \\ 
        OpenSeg\cite{ghiasi2022scaling}&ECCV22 &  ALIGN&Eff-B7\cite{pmlr-v97-tan19a}&  COCO Panoptic\cite{kirillov2019panoptic}+LOc. Narr.\cite{PontTuset_eccv2020}   & 8.8& 12.2& 28.6& 48.2 & 72.2 \\
        OVSeg\cite{Alpher12} &CVPR23 &  CLIP ViT-L/14&Swin-B&  COCO-Stuff+COCO Caption  & 9.0& 12.4& 29.6& 55.7 &  94.5 \\
        ODISE\cite{Alpher14}&CVPR23 & CLIP ViT-L/14&Stable Diffusion&  COCO Panoptic\cite{kirillov2019panoptic}   &11.1& 14.5& 29.9&   57.3&  -\\ 
        SAN\cite{Alpher17} &CVPR23 &  CLIP ViT-L/14&-&  COCO-Stuff  &13.7& 17.1& 33.3& 60.2 &  95.5 \\ 
        FC-CLIP\cite{Alpher18}&NeurIPS23 & ConvNeXt-L&-&  COCO Panoptic\cite{kirillov2019panoptic}   & 14.8& 18.2& 34.1&   58.4 &  95.4\\ 
        CAT-Seg\cite{Alpher19} &CVPR24& CLIP ViT-L/14&Swin-B&  COCO-Stuff   & 10.8& 20.4& 31.5&   62.0 &  96.6\\ 
        EBSeg\cite{Alpher32}&CVPR24 & CLIP ViT-L/14&SAM\cite{Alpher33}&  COCO-Stuff   & 13.7& 21.0& 32.8&   60.2 &  96.4\\ 
        SCAN\cite{liu2024open}&CVPR24 &  CLIP ViT-B/16&Swin-B&  COCO-Stuff  & 14.0& 16.7& 33.5& 59.3 &  \underline{97.2}\\
        SED\cite{Alpher20}&CVPR24 &   ConvNeXt-L&-&  COCO-Stuff  & 13.9& 22.6& 35.2& 60.6 &  96.1 \\
        USE\cite{wang2024use}&CVPR24 &   CLIP ViT-L/14&Swin-B,SAM\cite{Alpher33}&  COCO-Stuff  & 13.4& 15.0& \underline{37.1}& 58.0 &  - \\
        BBN\cite{pan2025purifythen}&TCSVT25 & CLIP ViT-L/14&-&  COCO-Stuff  & 14.2& \underline{23.5}& 34.5& \underline{63.7} &  96.8 \\
        GBA\cite{xu2024generalization}&TCSVT25 &      ConvNeXt-L&-&  COCO Panoptic\cite{kirillov2019panoptic}  & \underline{15.1}& 18.5& 35.9& 59.6 &  95.8 \\
        \rowcolor{gray!20} 
        \textbf{DCP-CLIP(Ours)}&- & CLIP ViT-L/14&-&  COCO-Stuff   & \textbf{16.4}& \textbf{24.3}& \textbf{38.1}&   \textbf{63.8} &  \textbf{97.3}\\ 
        \hline
    \end{tabular}
}
\vspace{1pt}
\label{table1}
\end{table*}

\begin{table*}[t]
\centering
\caption{\textbf{Efficiency comparison.} All results are measured on a single RTX 3090 GPU, with all methods using the base-scale vision-language model to ensure consistency in model size. We report Inference time and mlou on A-150 dataset in this table.}
\resizebox{\textwidth}{!}{
    \begin{tabular}{c|cccccc}
        \hline
         Method  &     \# of learnable params. (M)&\# of total params. (M) &     Training time (min) &     Inference time (s) &     Inference GFLOPs &     mIoU \\ 
        \hline
        ZSSeg\cite{xu2022simple} & 102.8& 530.8& 958.5&   2.73 &22,302.1&  20.5\\ 
        ZegFormer\cite{Ding_2022_CVPR}&103.3&  531.2   & 1,148.3& 2.70& 19,425.6&   16.9 \\ 
        OVSeg\cite{Alpher12}&  408.9   & 532.6& -& 2.00&   19,345.6 &24.8\\ 
        CAT-Seg\cite{Alpher19}&\textbf{70.3}  &\textbf{433.7}  & \underline{875.5}& 0.54& 2,121.1&   27.2  \\ 
        SED\cite{Alpher20}&204.8  &762.3  & 1302.4& \underline{0.21}& \underline{674.8}&   \underline{31.6}  \\ 
        \rowcolor{gray!20} 
        \textbf{DCP-CLIP(Ours)}&\underline{80.3}&  \underline{443.6}   & \textbf{496.4}& \textbf{0.20}& \textbf{616.2}&   \textbf{32.4} \\ 
        \hline

    \end{tabular}
}
\vspace{1pt}
\label{table2}
\end{table*}

\subsection{Experiment Setup}
\textbf{Datasets.}
Following previous open-vocabulary semantic segmentation methods \cite{Alpher12,Alpher18,Alpher32}, we train our model on the training set of COCO-Stuff \cite{caesar2018cocostuff}, and evaluate its generalization ability on several widely-used benchmarks, including ADE20K-150 \cite{zhou2019semantic}, ADE20K-847 \cite{zhou2019semantic}, Pascal Context-59 \cite{mottaghi2014role}, Pascal Context-459 \cite{mottaghi2014role}, and Pascal VOC \cite{everingham2010pascal}.

\textbf{COCO-Stuff} is an extended version of the original MS COCO dataset, which adds dense pixel-level annotations for both "thing" and "stuff" classes. It contains 164K images with semantic segmentation masks covering 171 categories, including 80 object (thing) classes and 91 background (stuff) classes. 

\textbf{ADE20K-150} is a widely-used semantic segmentation benchmark derived from the ADE20K dataset. It contains 150 semantic categories including a diverse set of objects and background regions across a wide variety of indoor and outdoor scenes. The dataset consists of 20,210 training images, 2,000 validation images, and 3,000 test images with densely annotated pixel-level labels.

\textbf{ADE20K-847} is an open-vocabulary semantic segmentation benchmark derived from the ADE20K dataset, containing 847 semantic categories. Unlike the standard ADE20K-150 which focuses on the 150 most frequent categories, ADE20K-847 includes a much larger and long-tailed vocabulary, covering rare and fine-grained concepts.

\textbf{Pascal Context-59} is a semantic segmentation benchmark consisting of approximately 5,000 images, with resolutions ranging from 240×320 to 516×775 pixels. Each image is densely annotated, assigning one of 59 semantic categories to every pixel.

\textbf{Pascal Context-459} shares the same set of images as Pascal Context-59, but includes a significantly larger vocabulary with 459 annotated semantic classes.

\textbf{Pascal VOC} is a widely-used benchmark for evaluating semantic segmentation models, containing 20 object classes with dense pixel-level annotations. It includes 1,464 training images and 1,449 validation images. In our experiments, we follow standard practice and use the validation set of Pascal VOC for evaluation.

\textbf{Evaluation Metric.}
For fair comparison, we follow prior works \cite{Alpher17,Alpher20} and adopt mean Intersection-over-Union (mIoU) as the evaluation metric, which is widely used in semantic segmentation. Specifically, mIoU measures the overlap between the predicted and ground truth segmentation masks by computing the ratio of their intersection to their union for each class, and then averaging the results across all classes.

\textbf{Implement details.}
We use the pretrained CLIP \cite{Alpher08} models released by OpenAI in our experiments, specifically the ViT-B/16 and ViT-L/14 variants. The model is trained using a per-pixel binary cross-entropy loss. We set the number of category templates P to 80. Our implementation is based on PyTorch \cite{paszke2019pytorch} and Detectron2 \cite{wu2019detectron2}. We adopt the AdamW optimizer \cite{loshchilov2019decoupled}, with a learning rate of $2\times 10^{-4}$ for our model and $2\times 10^{-6}$ for CLIP, and a weight decay of $10^{-4}$. The mini-batch size is set to 4. All models are trained on 4 NVIDIA RTX 3090 GPUs for 80K iterations.

\subsection{Comparisons With State-of-the-art Methods}
In Table ~\ref{table1}, we compare our proposed DCP-CLIP with some state-of-the-art open-vocabulary semantic segmentation methods. The table also list the corresponding vision-language model, feature extraction backbone, and training dataset employed by each method. Most approaches are built upon VLMs, with the exception of SPNet \cite{XianCVPR2019b} and ZS3Net \cite{Alpher05}. Some methods \cite{Alpher32,xu2022simple,Han_2023_ICCV} incorporate additional models, while others \cite{Alpher12,ghiasi2022scaling} utilize extra datasets or annotations. To ensure a fair comparison, the results are organized according to the underlying VLM used.

Under a similar VLM setup, our proposed DCP-CLIP consistently achieves superior performance across all five datasets compared to existing methods. Using CLIP ViT-B/16 as the vision-language backbone, our approach achieves mIoU gains of +0.9\%, +1.1\%, and +0.8\% on A-847, PC-459, A-150, respectively. Moreover, compared to OVSeg and OpenSeg, our DCP-CLIP does not require additional dataset or annotation. compared to SCAN and EBSeg, our DCP-CLIP does not require additional model. When leveraging the more powerful CLIP ViT-L, the performance improvements increase to +1.3\%, +0.8\%, and +1.0\% across the same datasets. These results demonstrate the effectiveness of the proposed DCP-CLIP framework.

In Table ~\ref{table2}, we further compare the efficiency of our method with recent approaches \cite{Alpher20,Alpher12,Ding_2022_CVPR}, evaluating key metrics including the number of learnable parameters, total parameter count, training time on COCO-Stuff, inference time, inference GFLOPs, and mIoU on the A-150 dataset. Our model exhibits strong efficiency in both training and inference stages.

\subsection{Ablation Study}
To gain a deeper understanding of the underlying behavior of our method, we present the results of a series of ablation studies in this section.

\begin{figure*}[t]
    \centering
    \includegraphics[width=\textwidth]{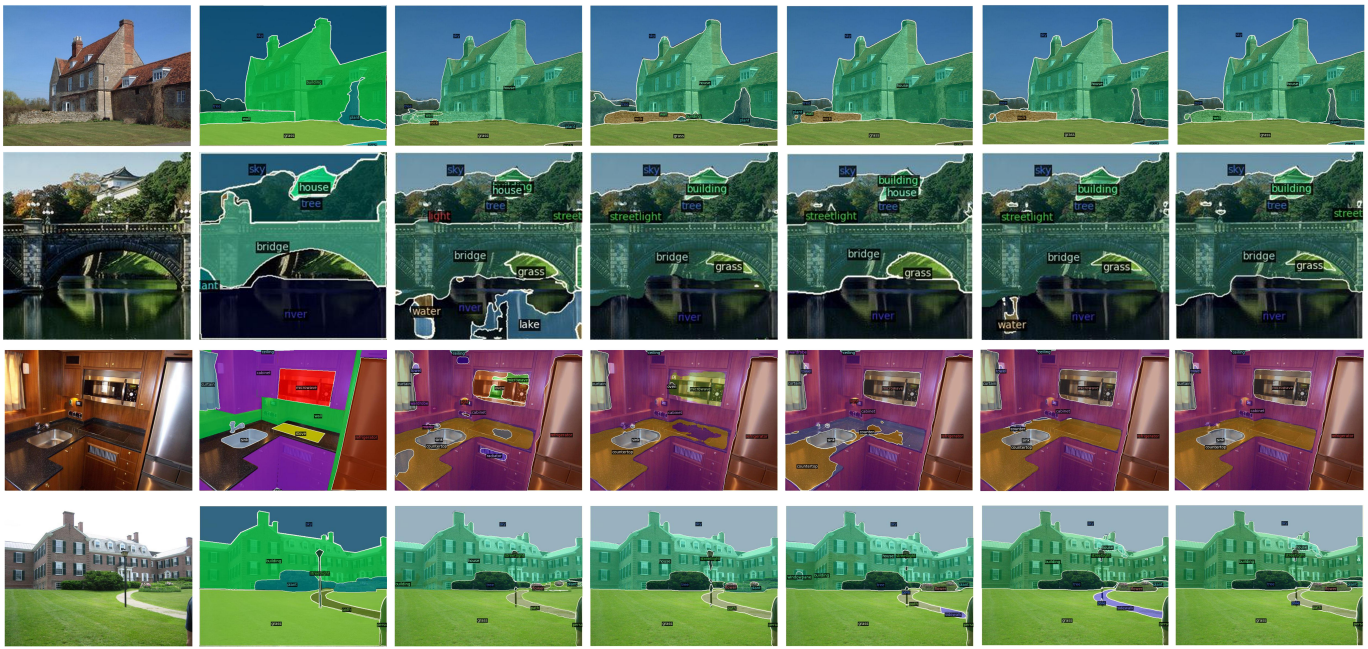}
\caption{\textbf{Visualization of ablation study results on open-vocabulary semantic segmentation.} From left to right are the input image, ground-truth segmentation, fine-tuning CLIP, and the predicted results after successively incorporating the DCS, TGA, SED, and TV modules. (Best viewed in color)}
    \label{fig5}
\end{figure*}

\textbf{Ablation Study of different components.} 
This section evaluates the contribution of each component to the improvement of mIoU on standard benchmarks dataset.We begin by establishing a baseline based on Fine-Tuning CLIP, DCS, TGA, SED, TRV are introduced in sequence. To further evaluate the individual contributions of the TGA and SED components, we conduct ablation studies by removing each component while keeping the rest intact. The corresponding results are presented in Table ~\ref{table3}. Overall, when the baseline is employed, we achieve the lowest mIoU scores of 7.1\%, 14.4\%, 23.1\%, 52.6\%, and 93.7\%on A-847, PC-459, A-150, PC-59 and PAS-20. Since PAS-20 shares most of its classes with the training datasets \cite{xu2023open}, the performance gap among all methods remains relatively small. In contrast, for more challenging datasets like A-847 and PC-459, the performance difference becomes considerably more pronounced. TIA and SED contribute individual gains of 3.1\%, 2.8\%, 6.7\% and 1.3\%, 1.9\%, and 2\% mIoU on A-847, PC-459 and A-150, respectively, underscoring the importance of both cross-modal and intra-modal feature learning. Finally, our TV further improves performance across all the benchmarks. Experimental results demonstrate that all proposed methods are effective for the open-vocabulary semantic segmentation task. To provide a more intuitive understanding of the module contributions, we present visualization results of the ablation study in Fig. 6, where the modules are progressively added to the framework. By observing the visualization results, we can see that the number of predicted categories decreases significantly after introducing the DCS module, which is most evident in the second row. When the TGA module is added to enable text–visual interaction, the segmentation quality improves progressively. Incorporating the SED module further refines the details, effectively recovering fine-grained structures. Notably, in the second and fourth rows, our model even segments streetlights that are not annotated in the ground truth but actually exist in the images. Finally, the TV module performs label verification, correcting misclassified regions and further enhancing the segmentation accuracy. For instance, in the second row, water is corrected to river, and in the fourth row, sidewalk is corrected to path, demonstrating the effectiveness of the TV module in refining category predictions.

\begin{table}[t]
\centering
\caption{\textbf{Ablation study for the contribution of each component of DCP-CLIP.} We conduct ablation study by gradually adding DCS, TGA, SED and TV to the Fine-Tuning CLIP baseline.}
\resizebox{\columnwidth}{!}{
    \begin{tabular}{c c c c|ccccc}
        \hline
        {DCS} & {TGA} &{SED} & {TV} & {A-847} &     {PC-459} &     {A-150} &     {PC-59} &     {PAS-20} \\ 
        \hline
         &  &  & & 7.1& 14.4& 23.1&  52.6 & 93.7 \\ 
        \checkmark & & & & 7.4& 14.3& 23.3&  52.9 & 93.6  \\ 
        \checkmark & \checkmark & & & 10.8& 17.7& 30.2&   56.3 &  94.6  \\ 
        \checkmark &  &\checkmark & &11.3& 18.6& 31.5&   57.1 &  94.5  \\ 
        \checkmark & \checkmark & \checkmark &&12.1& 19.6& 32.2&   57.9 &  95.7  \\ 
        \checkmark & \checkmark & \checkmark &\checkmark& 12.3& 19.7& 32.4&   58.1 &  95.9  \\ 
        \hline
    \end{tabular}
}
\vspace{1pt}
\label{table3}
\end{table}

\begin{table}[t]
\centering
\caption{\textbf{Ablation study for the text-guided alignment.} we gradually remove these components and evaluate their effects on model performance.}
\resizebox{\columnwidth}{!}{
    \begin{tabular}{c c |ccccc}
        \hline
        {CA} & {SA} & {A-847} &     {PC-459} &     {A-150} &     {PC-59} &     {PAS-20} \\ 
        \hline
         &  &   11.4& 18.5& 31.2&  57.4 & 94.4 \\ 
        \checkmark & &  12.1& 19.3& 31.9&  57.6 & 94.9  \\ 
         & \checkmark &  11.6& 18.7& 32.1&  57.9 & 95.5  \\ 
        \checkmark & \checkmark & 12.3& 19.7& 32.4&  58.1 &  95.9 \\ 
        \hline
    \end{tabular}
}
\vspace{1pt}
\label{table7}
\end{table}

\begin{table}[t]
\caption{\textbf{Ablation study for the structure of textual guidance.} We gradually ablate components of the proposed method to verify their effectiveness.}
\resizebox{\columnwidth}{!}{
\begin{tabular}{c|ccccc}
    \hline
    Method  &     {A-847} &     {PC-459} &  {A-150} &     {PC-59} &     {PAS-20} \\ 
    \hline
    full textual guidance (\uppercase\expandafter{\romannumeral1})& 12.3& 19.7& 32.4&   58.1 &  95.9 \\ 
    (\uppercase\expandafter{\romannumeral1}) w/o $F_{cls}\cdot E _{SC}$ (\uppercase\expandafter{\romannumeral2})& 12.2& 19.5& 32.5&   57.6& 95.5\\ 
    (\uppercase\expandafter{\romannumeral2}) w/o learnable prompt & 12.0& 19.1 & 32.0&  57.0& 95.1\\ 
    \hline
\end{tabular}
}
\vspace{1pt}
\label{table4}
\end{table}

\textbf{Ablation study for Text-guided alignment.} 
Table \ref{table7} shows the ablation results for the different attention branch in the text-guided alignment module. When both branches are removed, the model achieves the lowest performance, indicating that attention mechanisms are critical for effective alignment. Introducing only the CA branch improves results to 12.1\% on A-847 and 94.9\% on PAS-20, showing that CA effectively enhances cross-modal interactions between textual and visual features. Similarly, enabling only the SA branch yields 11.6\% on A-847 and 95.5\% on PAS-20, suggesting that context-aware fusion also contributes positively. The best overall performance is obtained when both CA and SA branches are jointly applied, reaching 12.3\% on A-847 and 95.9\% on PAS-20. These results confirm that the two attention branches are complementary:CA focuses on cross-modal alignment, while SA strengthens internal feature consistency—jointly leading to more robust text-guided representation learning. Table~\ref{table4} further presents the ablation results on different configurations of the proposed textual guidance in cross-attention branch. The full textual guidance ($I$) achieves the best performance on all datasets, confirming the effectiveness of our design. For example, it achieves 32.4\% on A-150 and 95.9\% on PAS-20. When removing the interaction term $F_{cls}\cdot E_{SC}$ ($II$), the performance drops consistently across all datasets. This indicates that modeling the semantic correlation between classification features and spatial context is essential for effective open-vocabulary generalization. For instance, mIoU decreases by 0.5\% on PAS-20 and by 0.3\% on PC-59. When the learnable prompt is further removed, the performance degrades more severely, especially on datasets with large-scale categories such as A-847 (-0.3\%) and PC-459 (-0.6\%), highlighting the importance of adapting textual representations dynamically. 

\begin{table}[t]
\caption{\textbf{Ablation study for Spatial Enhanced Decoder.} We evaluate the individual contributions of spatial enhancement, decoder layers, and feature fusion by modifying each component separately in (a), (b) and (c).}
\resizebox{\columnwidth}{!}{
    \begin{tabular}{c|c|ccccc}
        \hline
        \multirow{3}{*}{(a)}&Aggregation  &     {A-847} &     {PC-459} &     {A-150} &     {PC-59} &     {PAS-20} \\ 
        \cline{2-7}
        & None   &11.0& 18.1& 30.8&   56.6 &  94.7  \\ 
        & Spatial-enhance  & 12.3& 19.7& 32.4&   58.1 &  95.9 \\ 
        \hline
        \multirow{4}{*}{(b)}&Decoder Layer  &     {A-847} &     {PC-459} &     {A-150} &     {PC-59} &     {PAS-20} \\ 
        \cline{2-7}
        & 0 & 11.4& 18.8& 31.5&   57.0& 94.9 \\ 
        & 1   & 11.8& 19.4& 31.9&   57.6 & 95.2  \\ 
        & 2  & 12.3& 19.7& 32.4&   58.1 &  95.9 \\ 

        \hline
        \multirow{5}{*}{(c)}&Feature Fusion &     {A-847} &     {PC-459} &     {A-150} &     {PC-59} &     {PAS-20} \\ 
        \cline{2-7}
        & None & 11.6& 19.1& 31.3&   56.8 &  94.7 \\ 
        & $F_{patch}$   & 11.8& 19.4& 31.8&   57.6 & 95.5  \\ 
        & $F_{patch}+F_{sim}$  & 12.0& 19.5& 32.1&   57.8 & 95.7\\ 
        & $F_{patch}+F_{sim}+A_{CLIP}$  & 12.3& 19.7& 32.4&   58.1 &  95.9 \\ 
        \hline
    
    \end{tabular}
}
\vspace{1pt}
\label{table6}
\end{table}

\textbf{Ablation study for Spatial Enhanced Decoder.} 
Table~\ref{table6} presents a detailed ablation study evaluating the effectiveness of different components within the Spatial Enhanced Decoder. We analyze three key aspects: spatial-enhance aggregation, decoder layer depth, and multi-source feature fusion. The experiments are conducted across five open-vocabulary segmentation benchmarks to comprehensively assess generalization performance. In (a), we evaluate the impact of introducing the spatial enhancement mechanism. Compared to the baseline without aggregation, enabling this module consistently improves segmentation performance across all datasets. For example, the mIoU on A-150 improves from 28.2\% to 32.8\% (+4.6\%), and on PAS-20 from 93.6\% to 95.9\% (+2.3\%). This confirms that spatially enhanced contextual modeling can effectively strengthen semantic consistency and boundary awareness, especially for complex or fine-grained regions. In (b), we examine the influence of decoder depth. Results show a clear trend: increasing the number of decoder layers from 0 to 2 leads to steady improvements on all datasets. Notably, the mIoU on A-847 increases by 4.8\%, from 7.6\% to 12.4\%, while on PC-59 it improves from 52.9\% to 58.1\%. This demonstrates the necessity of deeper decoding structures to progressively refine coarse semantic representations into high-resolution predictions. In (c), we analyze different feature fusion strategies. Without fusion, the decoder relies solely on shallow features, resulting in sub-optimal performance. Introducing only $F_{\text{patch}}$ provides a noticeable boost. Further integrating $F_{\text{sim}}$ introduces similarity-aware representations that improve discriminability. Finally, adding $A_{\text{CLIP}}$ — the attention-guided alignment from the vision-language model — yields the best performance across all benchmarks. This indicates that multi-source fusion from diverse semantic levels and modalities can provide complementary cues and significantly enhance pixel-level prediction quality. Overall, these experiments verify the effectiveness and synergy of the spatial enhancement module, deep decoding structure, and multi-source fusion in building a robust and generalizable segmentation decoder under the open-vocabulary setting.

\begin{figure*}[t]
    \centering
    \includegraphics[width=\textwidth]{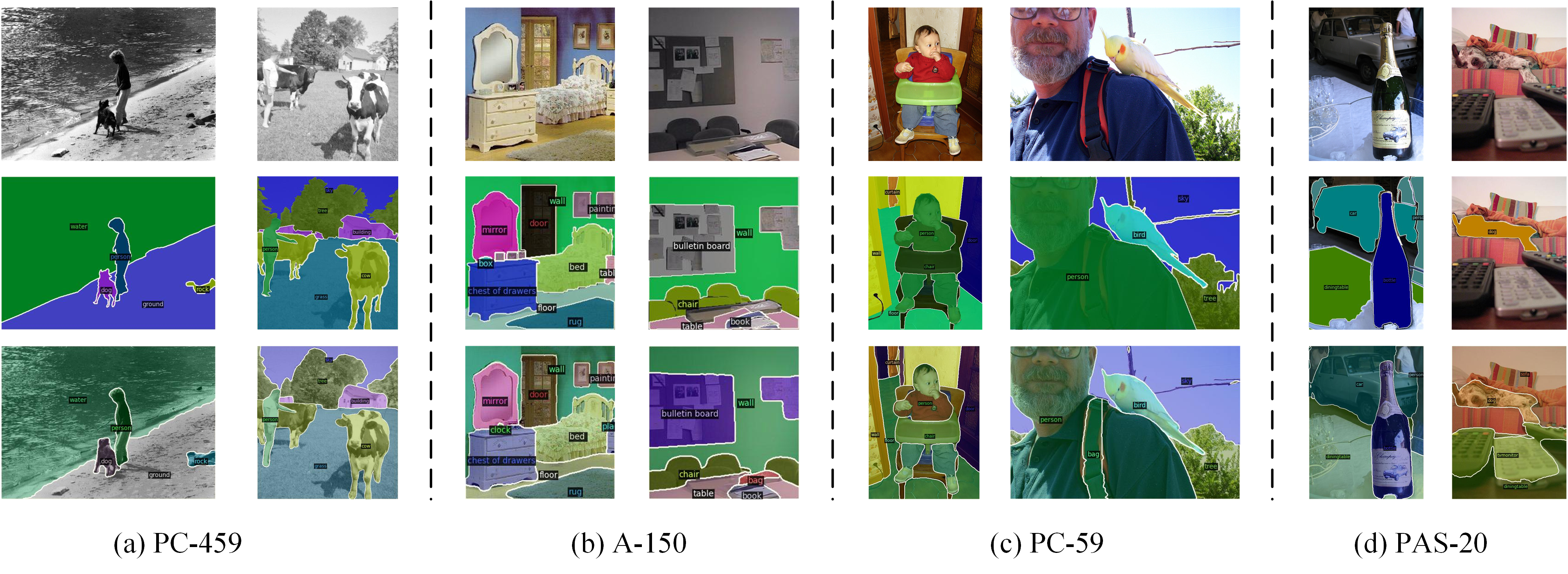}
    \caption{\textbf{Qualitative results on open-vocabulary semantic segmentation benchmarks.} Examples are shown from (a) PC-459, (b) A-150, (c) PC-59, and (d) PAS-20 datasets. From top to bottom are the input image, ground-truth segmentation, and our model’s prediction. (Best viewed in color)}
    \label{fig6}
\end{figure*}

\textbf{Qualitative results.} 
Figure ~\ref{fig6} presents some qualitative results on open-vocabulary semantic segmentation benchmark. We display the input images, corresponding ground truth, and our model's predictions to demonstrate its performance across diverse and challenging scenarios. Our segmentation results not only capture foreground objects in seen classes such as "person", "dog", "cow", and “bed", but also successfully identify unseen classes like "bulletin board" and "chest of drawers", as illustrated in the (a) and (b). The predictions not only capture fine-grained object boundaries but also reveal small objects that are unannotated in the ground truth, such as the “bag” in the second column of (c), the “door” in the upper-left of first column of (c) and the "TV monitor" in the second column of (d). These results highlight the model’s strong open-vocabulary generalization and ability to infer meaningful object boundaries beyond the training categories.

\begin{table}[t]
\caption{\textbf{Analysis of threshold of dynamic category selection.} Here, We report the inference time on a single RTX 3090 GPU, evaluated on the A-847, A-150 and PAS-20 dataset. inference time is evaluated on A-150 dataset.}
\resizebox{\columnwidth}{!}{
    \begin{tabular}{c|ccc|ccc}
        \hline
        \multirow{2}{*}{$\theta$}  & \multirow{2}{*}{A-847} & \multirow{2}{*}{A-150}& \multirow{2}{*}{PAS-20} &  {Memory} &    {Training time }&inference time\\ 
         &    & &  & $(GiB)$ & $(min)$ & $(s)$\\
        \hline
        0.9  &   11.9 &32.0& 96.0  & 12.1& 461.4&0.19\\ 
        0.8   &   12.1 & 32.2&96.1  & 12.6&478.9&0.19\\ 
        0.5  &   12.3 & 32.4& 95.9  & 13.2& 496.4&0.20\\
        0.2  & 12.4& 32.6& 95.6 & 15.5& 528.6&0.22\\ 
        0 &   12.4 & 32.5 & 95.2  & 20.1& 869.5&0.51\\ 
        \hline
    
    \end{tabular}
}
\vspace{1pt}
\label{table5}
\end{table}

\subsection{Analysis of Parameter Settings}

\textbf{Effect of Confidence Threshold.} 
Table~\ref{table5} presents a comprehensive evaluation of the proposed dynamic category selection mechanism under different threshold values $\theta$. As $\theta$ increases, fewer categories are retained during training and inference, resulting in a significant reduction in computational overhead. Specifically, at $\theta=0$, where all categories are included, the model achieves 32.5 \% on A-150 and 95.2 \% on PAS-20, but at the cost of substantial resource consumption—requiring 20.1GiB of memory, 869.5 minutes of training time, and 0.51 seconds per inference. As $\theta$ increases, the number of selected categories gradually decreases. Consequently, memory usage, training time, and inference time are significantly reduced without notable performance degradation. Notably, when $\theta=0.9$, the model still maintains competitive performance with 32.0 \% on A-150 and 96.0 \% on PAS-20, while cutting memory usage down to 12.1GiB, training time to 461.4 minutes, and inference time to just 0.19 seconds. This demonstrates that our dynamic category selection mechanism can substantially improve computational efficiency with minimal accuracy loss. 

\section{Conclusion}
In conclusion, We  propose DCP-CLIP, a novel coarse-to-fine framework with dual interaction that explicitly enhances communications between textual and visual spaces. Our method enhances cross-modal communication by introducing a dual interaction between textual and visual spaces, while incorporating a dynamic category selection mechanism to improve computational efficiency. Extensive benchmark results validate the state-of-the-art performance of our method. Moreover, the efficiency comparison demonstrate our dynamical category select mechanism provides an efficient solution to improve computational efficiency for open-vocabulary semantic segmentation. 

{
\bibliographystyle{IEEEtran}
\bibliography{ref}
}
\vfill
\end{document}